\documentclass[runningheads]{llncs}
\usepackage{graphicx}
\usepackage{subfigure}
\usepackage{multirow}
\usepackage{amssymb}
\usepackage{hyperref}
\hypersetup{hypertex=true,
colorlinks=true,
linkcolor=blue,
anchorcolor=blue,
citecolor=blue}

\begin{document}
\title{Multi-View Stereo Network with Attention Thin Volume
% \thanks{Supported by organization x.}
}
%
%\titlerunning{Abbreviated paper title}
% If the paper title is too long for the running head, you can set
% an abbreviated paper title here
%
% \author{WZH\inst{1} \and
% MZP\and
% XC\inst{1,2,3}\and
% HJ\inst{1,3}\and
% XJ\inst{1,4}}\and
% CJT
\author{
Zihang Wan}
% %
% \authorrunning{F. Author et al.}
% % First names are abbreviated in the running head.
% % If there are more than two authors, 'et al.' is used.
% %
% \institute{College of Intelligence and Computing, Tianjin University, Tianjin, 300350, China 
% \and
% Higher Research Institute(shenzhen), University of Electronic Science and technology of China, Shenzhen, 518063, China
% \and
% The Hong Kong Polytechnic University, Hong Kong
% \and
% Tianjin University of Traditional Chinese Medicine,Tianjin, 300193, China\\
% }
%
\maketitle              % typeset the header of the contribution
\begin{abstract}
We propose an efficient multi-view stereo (MVS) network for inferring depth value from multiple RGB images. Recent studies use the cost volume to encode the matching correspondence between different views, but this structure can still be optimized from the perspective of image features. First of all, to fully aggregate the dominant interrelationship from input images, we introduce a self-attention mechanism to our feature extractor, which can accurately model long-range dependencies between adjacent pixels. Secondly, to unify the extracted feature maps into the MVS problem, we further design an efficient feature-wise loss function, which constrains the corresponding feature vectors more spatially distinctive during training. The robustness and accuracy of the reconstructed point cloud are improved by enhancing the reliability of correspondence matches. Finally, to reduce the extra memory burden caused by the above methods, we follow the coarse to fine strategy. The group-wise correlation and uncertainty estimates are combined to construct a lightweight cost volume. This can improve the efficiency and generalization performance of the network while ensuring the reconstruction effect. We further combine the previous steps to get what we called attention thin volume. Quantitative and qualitative experiments are presented to demonstrate the performance of our model.

\end{abstract}
\section{Introduction}
Multi-view stereo is dedicated to recovering the depth value from overlapping photo sets, which has become a key issue in computer vision. In the past, many traditional methods adopt manually designed feature descriptors to determine the correspondence between pixels in different pictures. They often apply engineering regularization to recover three-dimensional point clouds. However, these methods have poor reconstruction effects in smooth areas such as weak textures and specular reflections. With the introduction of deep learning, the cost volume is employed for simulation matching, and the quality of the reconstruction has been greatly improved. 

Referring to the previous excellent learning-based MVS work, it is not difficult to find using cost volume to complete correspondence matching is a crucial step in the whole pipeline. We focus on the construction of cost volume, start from image features, and propose an attention-thin volume structure to better solve the MVS problem.

For cost volume aggregating, the first step is to extract the feature maps from input images. In MVS task, the depth value of adjacent pixels changes either very drastically (at the edge of the object contour) or very smoothly (at the continuous surface of the object). This requires paying attention to the interrelationship between each pixel and its nearby pixels, that is, the long-range dependence in a single small context. CNN or global attention mechanisms~\cite{LANet} are not suitable enough, and the self-attention~\cite{stand-alone} has a lot of potential here.

At the same time, based on richer feature information, subsequent feature matching is also crucial to the construction result. Ideally, the feature vectors corresponding to the same pixel in different views should have sufficient similarity. In this way, robustness and accuracy can be ensured when using the cost volume to match the corresponding relationship. Therefore, we design a feature-wise loss function to strengthen this constraint, which can be shown to work well through experiments.

Besides, In the process of aggregating feature maps into cost volumes, previous works~\cite{Cas-MVSNet,UCSNet} have proved that direct variance aggregation will generate redundant information, and the redundancy of depth sampling will also lead to memory consumption. Therefore, we follow the coarse to fine strategy, specifically, combining group-wise correlation and uncertainty estimate. The similarity between feature map channels is calculated first and then aggregated into cost volumes, which reduces memory usage while improving information utilization efficiency. As for the coarse to fine framework, we use the uncertainty estimate strategy~\cite{UCSNet} to reduce the number of depth samples. These works make the similarity measurement of the corresponding regions of different pictures in the cost volume more accurate and efficient.

Our main contributions are as follows:
\begin{enumerate}
\item We design a multi-scale feature extractor with a self-attention mechanism to fully extract the interrelationship between adjacent pixels, so richer information can be provided for subsequent matching steps.
\item We design a novel feature-wise loss function to constrain the feature maps extracted by our feature extractor, which encourages the enhancement of features with spatial consistency. This improves the robustness and accuracy of matching correspondences.
\item We organically combine the group-wise correlation and uncertainty estimation strategies to complete the construction of the cost volume. The network is iteratively refined between different scales, reducing the memory burden of the framework, and optimizing the use of feature information.
\item We achieve good qualitative results on benchmark dataset. The complete performance is improved by 9.7 $\%$ on the basis of the backbone, and the overall is improved by 3.7 $\%$.
\end{enumerate}
\section{Related Wrok}
This chapter summarizes the related work of learning-based multi-view stereo methods and introduces several research directions that our work also covers.
\subsection{Learned MVS} 
Recently, learning-based methods have shown remarkable performance on multi-view stereo. SurfaceNet~\cite{SurfaceNet} predicts voxel confidence to determine if the pixel is on the surface and reconstructs the 2D surface of the scene. The concept defined by DeepMVS~\cite{DeepMVS} and MVSNet~\cite{MVSNet} is the basis of many similar works. Among them, the key idea is to perform plane sweep through homography, construct cost volume for correspondence matching based on variance, and use 3D CNN for depth map regression. RMVSNet~\cite{RMVSNet} introduces a recurrent neural network and uses Gate Recurrent Unit(GRU) to replace the 3D CNN network, so it greatly optimizes the memory consumption. MVSCRF~\cite{MVSCRF} introduces Conditional Random Field (CRF) for depth map prediction, and CRF is implemented in a cyclic neural network to facilitate end-to-end training. Point-MVSNet~\cite{Point-MVSNet} proposes to use a smaller cost volume to first generate a rough depth map, and then use an iterative refinement network based on point clouds to further optimize the previously obtained depth map. PMVSNet~\cite{PMVSNet} learns the patch-wise confidence from the reference image and the source images, and adaptively performs cost volume aggregation. PVA-MVSNet~\cite{PVA-MVSNet} introduces two novel adaptive view aggregations: pixel-level view aggregation and voxel-level view aggregation, which combine the cost variances in different views with small extra memory consumption. 
\subsection{Coarse To Fine} 
Cost volume requires a large number of depth hypotheses to cover potential depth range. Limited by the memory bottleneck, it is difficult to reconstruct a high-resolution depth map. The strategy of coarse to fine is introduced to solve this problem by iterative descaling depth sampling. CasMVSNet~\cite{Cas-MVSNet} proposes that by constructing cascade cost volume and reducing the number of depth hypothesis planes scale by scale, the network can more flexibly adapt to the needs of large-scale reconstruction scenarios. CVP-MVSNet~\cite{CVP-MVSNet} continuously improves the reconstruction accuracy between different scales by learning deep residuals. UCSNet~\cite{UCSNet} proposes to use uncertainty estimation to construct adaptive thin volumes (ATV), which can replace the previous plane sweep volume so that the hypothetical depth plane can become a curved continuous one. This strategy further improves the completeness and accuracy of the MVSNet framework.
\subsection{Attention Mechanism} 
With the success of the attention mechanism in other fields of computer vision, this method has also been introduced to MVS by many researchers. Att-MVS~\cite{Att-MVS} proposes enhanced matching confidence based on the attention mechanism, which combines the extracted pixel-level matching confidence with the context information of the local scene to improve matching robustness. At the same time, a regularization module guided by the attention mechanism was developed to aggregate and regularize the matching confidence to obtain a more accurate probability volume. PA-MVSNet~\cite{PA-MVSNet} uses a multi-scale feature pyramid with an attention mechanism to capture larger receptive fields and richer information. The results of the pyramid attention modules of different scales are directly used for the next scale prediction. Compared with previous algorithms, the accuracy has been greatly improved. LA-Net~\cite{la-net} propose
% focused on two shortcomings of previous methods, one is that they ignore the dependence between pixels, and the other is the ineffectiveness of weak texture or occlusion area matching. 
a remote attention network to selectively aggregate reference features to each location to capture the remote interdependence of the entire space.
\begin{figure}[h]%
\centering
\includegraphics[width=1\textwidth]{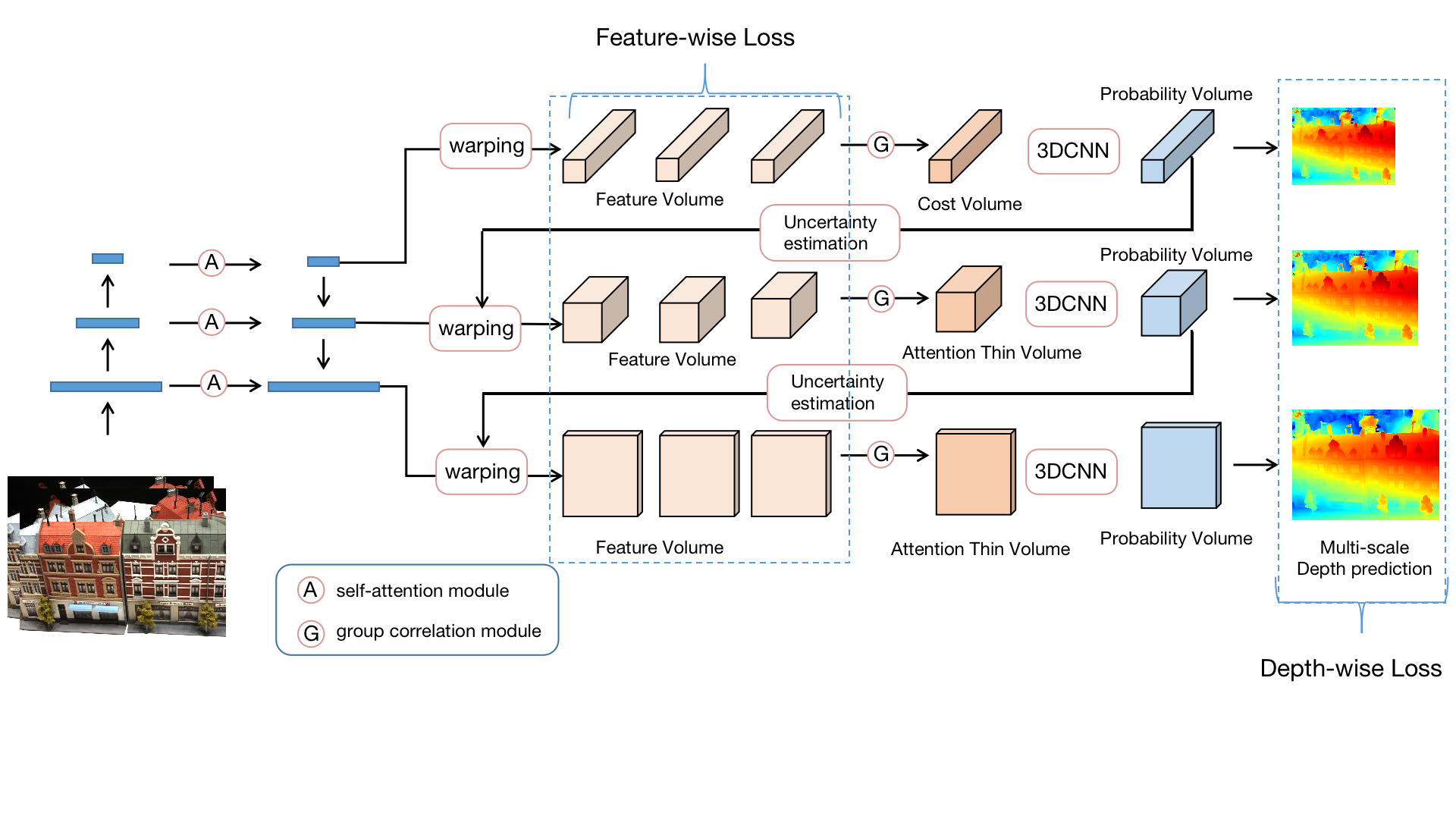}
\caption{Network architecture of our proposed network. The network is divided into three scales, and the image resolution increases sequentially from top to bottom. As shown in the figure, the original cost volume is constructed on the coarsest scale, and at the finer two scales, attention thin volume is constructed to express the correspondence between different views.}\label{architecture}
\end{figure}
\section{Method}
% To fully utilize the existing framework to capture image information and improve the quality of 3D reconstruction results, we designed an efficient system according to the characteristics of the MVS task.
This section describes our process for depth inference. Given a reference image $I_{0}\in R^{H \times W}$ and N source images $\{ I_{i} \}  _{i=1}^N \in  R^{H \times W}$. The camera intrinsic, rotation matrix, and translation vector $ \{ K_{i}$,$  R_{i}$,$   t_{i} \} _{i=0}^N $ for each input view i are known. Our goal is to predict the depth map $D_{0}$ corresponding to reference image $I_{0}$. 

The overall architecture is shown in Figure~\ref{architecture}. In Section \ref{3.1}, we introduce our designed feature extractor, and in Section \ref{3.2}, we show how we fuse the group-wise correlation into coarse-to-fine framework, specially how to aggregate the feature maps into attention thin volume, Section \ref{3.3} introduces the depth map regression, Section \ref{3.4} introduces the our designed feature-wise loss function.
\subsection{Multi-scale Feature Extractor With Self-attention Mechanism}\label{3.1}
Previous learning-based MVS methods mostly used multiple down-sampled convolutional layers or 2D UNet for feature extraction. For our consideration of neighborhood pixel association and information fusion between multi-scale images, we combine UNet~\cite{UNet} and self-attention mechanism~\cite{stand-alone,saUNet,aacvp} to design a multi-scale feature extractor with self-attention mechanism. 
% The architecture of our feature extractor is shown in Figure~\ref{saUNet}.
% \begin{figure}[h]%
% \centering
% \includegraphics[width=1\textwidth]{imgs/saUNet.png}
% \caption{The architecture of our Multi-scale feature extractor with self-attention mechanism}\label{saUNet}
% \end{figure}
\subsubsection{Multi-scale feature}
Multi-scale is implemented by U-Net structure with encoder and decoder. Output at three different scales, preserving low-level and high-level semantic information. In the skip-links process, we employ a self-attention mechanism module to capture local feature interrelationships. 
In this paper, we use $s \in (0,1,…,n)$ to represent the different image scales from original to rough, and the corresponding image resolutions in different scales are represented as $H_{s} , W_{s} =  H  \times  {1/2}^{s}  , W \times  {1/2}^{s}$, In this way, for each input image, feature maps of different scales $F_{i}^s \in  R^{ H_{s}  \times  W_{s}  \times C}$ can be obtained, and C is the number of channels of the output feature map, thus this structure can be adapted to the subsequent coarse to fine strategy.
\subsubsection{Self-attention mechanism}
The self-attention mechanism is mainly used to capture neighborhood pixel dependencies, and the relative positions\cite{relative-position} between pixels are used. Given the network weight $W \in  R^{k \times k \times  d_{in} \times  d_{out}  }$, the output of each pixel position (i, j) after convolution is denoted as:
\begin{equation}
  y_{i,j} = \sum\limits_{a,b \in B} Softmax_{ab}  ( q_{ij}  ^{T} k_{ab}+ q_{ij}  ^{T}  r_{a-i,b-j})  v_{ab} 
\end{equation}
where B represents the image block, which size is as same as the kernel, $q_{ij} =  W_{q}  \times  x_{ij}$, $ k_{a b} =  W_{k}  \times  x_{ab}$, $v_{ab} =  W_{v}  \times  x_{ab} $ represent queries, keys and values respectively, $r_{a-i,b-j}$ indicates the offset of pixel rows and columns, W are the learned parameters of each component. The overall calculation of the formula is divided into four steps:
\begin{enumerate}
\item Determine the pixel position (i, j), the pixel area image block B(a, b) and the pixel relative offset $r_{a-i,b-j} $.
\item Calculate queries $q_{ij}$, keys $k_{a b}$, and values $v_{ab}$.
\item Use inner product to calculate the similar relationship between queries and keys and pixel offset, then map the result via softmax.
\item Perform the final calculation on the values and the similarity in step 3 to obtain the weight evaluation result of the pixel.
% \item Traverse each valid pixel in the entire image and repeat steps 1 to 4 to obtain a feature map with self-attention mechanism information.
\end{enumerate}
The output $y_{i,j} $ represents the similarity between query and the elements in image block B, and at the same time, it enjoys translation equivariance. At each scale s, given an input image $I_{i} $, a feature map $F_{i}^s \in  R^{ H_{s}  \times  W_{s}  \times C} $ that aggregates long range dependencies can be obtained by our feature extractor. 
\subsection{Attention Thin Volume Generation}\label{3.2}
% Learning based multi view stereo measures the similarity between corresponding image blocks and determines whether they match by constructing a 3D cost volume. As aforementioned, previous aggregate method\cite{MVSNet,Cas-MVSNet} is just for warping different feature maps into one independent cost volume, and then performing a regular regression. This leads to insufficient use of the neighborhood channel information contained in different feature maps, and consumes huge computational resources. 
To perform a more efficient similarity measurement, we follow the framework of coarse to fine, and introduce group-wise correlation\cite{group-wise} based on uncertainty estimate\cite{UCSNet}, thereby constructing what we called attention thin volume, which reduced memory burden while minimizing information loss. 
\subsubsection{Group-wise correlation}
At each scale, after obtaining the feature map, we warp source image feature maps to each hypothetical depth plane upon the reference camera frustum, and aggregate them to a cost volume by group-wise correlation. This can provide a wealth of similarity measurement, while reducing the memory burden. The calculation steps are as follows:
\begin{enumerate}
\item Homography is used to warp the feature maps of source images $F_{1}^s,F_{2}^s,…,F_{N}^s $ to the depth hypothesis  $d_{i}  \in [ d_{min}, d_{max}  ],i  \in   [0,D_{s}-1]$ upon the reference camera frustum, where $D_{s}$ is different at each scales,
% It is a fixed value on the coarsest scale, and then can be adaptively adjusted according to the depth inference results of the upper scale, 
$d_{i}$ is uniformly sampled within the range $ [d_{min}, d_{max}] $. The homography between two feature maps can be expressed by the following formula:
\begin{equation}
  H_{i}(d) =  K_{i}  T_{i}   T_{0} ^{-1}  K_{0} ^{-1} 
\end{equation}
where $ \{K_{i} , T_{i} \} $ is the camera intrinsic and extrinsic corresponding to different images. 
% When a pixel in source image warped to the pixel in the reference image $I_{0}$ at location $(x,y)$ and depth d. Multiply $H_{i}(d)$ by the homogeneous vector $(x_{d},y_{d},d,1)$ to find the homogeneous coordinates of the corresponding pixel position in each $I_{i}$.
\item After homography transformation, at each depth hypothesis d, there corresponds to N+1 feature maps $F_{0,d}^s,F_{1,d}^s,…,F_{N,d}^s \in  R^{ H_{s}  \times W_{s} \times C} $. Divide each feature map into G groups according to the feature channel, and then use the inner product to calculate the similarity of different sub-feature maps in the same group $g \in [0,G-1] $. The calculation formula is
\begin{equation}
  M_{d,i} ^{g,s} =  \frac{1}{C/G}  \langle F_{0,d}^{g,s},F_{i,d}^{g,s}   \rangle
\end{equation}
where $\langle ·,· \rangle$represents the inner product, $F_{i,d}^{g,s} $ is the g-th group of sub-feature maps belong to the i-th feature map at depth d. $M_{d,i}^{g,s}  \in R^{1 \times  H_{s}  \times  W_{s} }$ are the similarity measurement of the corresponding two ones. After that, $M_{d,i}^{g,s} $ are aggregated in the order of group and depth, and the feature volume corresponding to  the i-th source image can be obtained, denoted as $V_{i}^{s}  = concat( M_{d,i} ^{g,s}), V_{i}^{s} \in  R^{G \times  H_{s} \times  W_{s} \times  D_{s}}$.
\item Variance aggregation is performed on the feature volume corresponding to the feature maps of different views
\begin{equation}
    C^{s} = \frac{1}{N} \sum\limits_{i=1}^{N} (V_{i}^s-\overline{ V_{i}^{s} } )^2
\end{equation}
where, $ \overline{ V_{i}^{s} } $ is the mean volume of different feature volumes, the variance calculation can allow any number of input images.
\end{enumerate}
It is worth noting that the group-wise correlation strategy will be applied at all scales, while at the last two scales with the higher resolution, the uncertainty estimate strategy will play an important role in determining the depth hypothesis.
\subsubsection{Uncertainty estimate}
% The core idea of coarse to fine framework is that the depth sampling range of a certain scale is determined by the prediction result of the previous scale. Here we 
We follow the idea of uncertainty estimate~\cite{UCSNet} to dynamically adjust the depth hypothesis interval. So the uniformly defined sweep plane becomes a curved continuous variable one.
% that is, the current-scale predicted depth value of a certain pixel and the relative variance of each depth value in the depth hypothesis interval are used to determine the hypothetical depth interval of the pixel in the next scale.\\
Specifically, the depth assumption at the coarsest level $L_{1}(x)$is fixed based on the results of the sparse reconstruction. Then, on the basis of the probability volume $P_{s}$ generated at each scale, we calculate the variance of the probability distribution
\begin{equation}
   U_{s} (x)=  \sum\limits_{d=1}^{D_{s}}  P_{s,d}(x)  \cdot (L_{s,d}(x)-D^{s}(x)) ^2
\end{equation}
and the corresponding standard deviation is $\widehat{ \sigma (x)}= \sqrt{U_{s}(x)}$. So we set the depth prediction range of the next scale to 
\begin{equation}
   DR_{(s+1)}(x)  = [D^{s}(x)- \lambda  \widehat{ \sigma (x)}, D^{s}(x)+ \lambda  \widehat{ \sigma (x)}]
   \label{cof:inventoryflow}
\end{equation}
where, $\lambda$ is a scale parameter, which can adjust the range of the depth hypothesis range. Sampling a number of $D_{(s+1)}$ depth values uniformly on $DR_{(s+1)}(x)$ can obtain the corresponding $L_{(s+1)}(x)$. 
Thus, the cost of correspondence matching between the reference view and all source views under each depth hypothesis can been encoded into the attention thin volume.
% %At this point, the core of the adaptive coarse to fine strategy has been described.
% We can get our attention thin volume upon the above steps. 
Our attention thin volume can integrate more crucial information by enhancing inter-channel connections and reduce memory burden. 
% From the coarsest scale with the lowest resolution to the finest scale, the corresponding predicted depth map $D^{s}  \in R^{H_{s} \times W_{s} }$ will be output, and the depth map of the last scale $D =  D^{s_{max}}$ is the final prediction result of the network
\subsection{Depth Map Regression}\label{3.3}
At each scale, after the aggregated cost volume(or attention thin volume) is obtained, a softmax-based regularization network is applied to get probability volume $P_{s}   \in R^{ H_{s}  \times  W_{s} \times  D_{s}  }$. It is treated as the weight of depth hypotheses. And we use formula
\begin{equation}
    D^{s}(x) = \sum\limits_{d=1}^{ D_{s} } P_{s,d}(x)  \cdot L_{s,d}(x)
\end{equation}
to calculate the final depth prediction. $L_{s,d}(x)$ represents the value of each depth hypothesis, and $P_{s,d}(x)$ is the probability that the pixel x corresponds to the depth value $L_{s,d}(x)$.

\subsection{Loss Function}\label{3.4}
We design a multi-metric loss function to constrain network training at different stages of the MVS task, which consists of feature-wise loss function and the traditional depth map loss function. The multi-metric loss function $L$ is the sum of the above two ones.
\begin{equation}
    L = \sum (\beta_{1} L_{fea} + \beta_{2} L_{depth})
    \label{mm-loss}
\end{equation}
\subsubsection{Feature-wise loss function}
As an information source, the feature map will directly affect the matching relationship in the cost volume, ideally, feature vector of the same pixel at different views should have strong spatial invariance. When the camera intrinsic and extrinsic are known, it is easy to warp the feature vector of a special pixel in source images to the reference image view. 
To enhance the matching of corresponding pixels during cost volume aggregation, we use \textbf{the position term} to encourage degenerating the features with weak invariance. The formula is expressed as:
\begin{equation}
   L_{pos} = \sum\limits_{i,j \in V} \Vert F(T_{i})-F(T_{j})\Vert_{2}
\end{equation}
where, V is the valid area in the reference image, $T$ denotes an input tensor and $T_{i},T_{j}$ the corresponding pixel feature vector pairs between two images, F represents the transform by our multi-scale feature extractor. At the same time, to reduce the degradation of reconstruction completeness due to the weakening of low-consistency features, we design a \textbf{neighbor balance loss} as a fault tolerance term to appropriately enhance feature similarity within the neighborhood. The formula is expressed as
\begin{equation}
   L_{nei} = \sum\limits_{i,j \in B} \Vert F(T_{i})-F(T_{j})\Vert_{2}
\end{equation}
where, B is the image block centered on i, j represents the other pixels in B. And we set the block size to 3*3 in our most experiments. The complete feature-wise loss is expressed as 
\begin{equation}
   L_{fea} = L_{pos}+\epsilon L_{nei}
\end{equation}
$\epsilon$ is a scale parameter, which is set to 0.01 in our work.
\subsubsection{Depth map loss function}
For the depth map predicted by the network, we follow the classic approach of the past and apply supervision to all outputs. The weighted sum of the l1 norm between the depth map prediction and the ground truth are used to construct the depth map loss function
\begin{equation}
   L_{depth} =  \sum\limits_{s=1}^{s_{max}}  (\Lambda^s \cdot \sum\limits_{x \in \Omega } ||D^s(x)-D_{GT}^s(x) || _1 )
   \label{dm-loss}
\end{equation}
where, $\Lambda^s$ is the loss weight of scale s , $D_{GT}^s$ is the ground truth of the depth map adapted to the resolution of scale s, and $\Omega$ is all the valid pixels in the image. 

\section{Experiments}\label{sec4}
In this section, we evaluate our method on the benchmark dataset, which proves the effectiveness of our proposed network. 
% We briefly introduce two benchmark datasets and the corresponding evaluation criteria in Section \ref{4.1}, then we describe the implementation details in Section \ref{4.2}, and finally we show the experimental results in Section \ref{4.3} and ablation studies in Section \ref{4.4}.
\subsection{Datasets and evaluation metrics}\label{4.1}
The effect of our work is evaluated on two benchmark datasets, one is the indoor DTU dataset~\cite{DTU}, the other is the outdoor Tank\&Temples dataset~\cite{Tank}.
\subsubsection{DTU dataset} contains 124 different indoor scenes, using an industrial robot arm equipped with a structured light scanner to scan structured light for different objects. Each scene contains 49 or 64 images which the viewpoints and lighting conditions are all deliberately designed. Camera pose parameters and point cloud ground truth have been provided. The image resolution is 1600*1184, and the depth range of each scene is between 425mm to 935mm. 
\subsubsection{Tank\&Temples dataset} contains two scene sets, namely intermediate and advanced. We use the intermediate set for evaluation. There are 8 different outdoor scenes, namely Family, Francis, Horse, Lighthouse, M60, Panther, Playground, and Train. The dataset is very challenging, the reconstruction scene is very large, and there are many reflections and occlusion on the surface of the object.
\subsubsection{Evaluation metrics} Consistent with previous methods, the accuracy and completeness of the distance measurement are used for evaluation of the DTU dataset. The accuracy and completeness of the percentage measurement are used for the Tanks\&Temples dataset. To obtain a summary measure of accuracy and completeness, the distance measure uses their mean value, and the percentage measure uses \textit{$F_{1}$} scores.
\subsection{Implementation}\label{4.2}
\begin{table}[h]\scriptsize%
\caption{ Performance comparison on DTU dataset (lower means better).} \label{DTU_results}
\begin{center}
% \resizebox{\textwidth}{}{
\begin{tabular}{llll}
\hline
method & Acc. & Comp. & Overall\\
\hline
Camp	&0.835	&0.554	&0.695\\
Furu	&0.613	&0.941	&0.777\\
Tola	&0.342	&1.190	&0.766\\
Gipuma	&\textbf{0.283}	&0.873	&0.578\\
\hline
MVSNet	&0.396	&0.527	&0.462\\
R-MVSNet	&0.383	&0.452	&0.417\\
Vis-MVSNet	&0.369	&0.361	&0.365\\
Cas-MVSNet	&0.346	&0.351	&0.348\\
CVP-MVSNet	&0.296	&0.406	&0.351\\
UCSNet	&0.338	&0.349	&0.344\\
\hline
Att-MVS	&0.383	&0.329	&0.356\\
LANet	&0.320	&0.349	&0.335\\
AACVP-MVSNet	&0.357	&0.326	&0.341\\
\hline
ours &0.347 &\textbf{0.315} &\textbf{0.331}\\
\hline
\end{tabular}
\end{center}
\end{table}
\subsubsection{Post processing}
We fuse all the predicted depth maps into a dense point cloud using the same post processing method~\cite{fusible} as our baseline~\cite{UCSNet}, which consist three main steps: photo-metric filtering, geometric consistency filtering, and depth fusion.
\subsubsection{Training details}
We train our proposed network on DTU training set. As same as previous works, we perform Poisson Surface Reconstruction~\cite{Poisson} to get the ground truth depth map. For each reference image, we use 2 source images during training, which resolution are $H \times W=640 \times 512$. The number of depth samples in the 3 scales are $D_{1}=32,D_{2}=16,D_{3}=8$, the init depth sampling range is $[425mm,935mm ]$ marked in the dataset. The scale parameter in the formula \ref{cof:inventoryflow} is set to $\lambda = 1.5$. The ratio of the two loss functions in formula \ref{mm-loss} is $\beta_{1}=1,\beta_{2}=1$. The weight of loss at different scales in formula \ref{dm-loss} is $\Lambda_s=\{0.5,1,2 \}$. The number of group is 8. We train the network for 16 epochs on a Tesla V100 GPU. We use the Adam optimizer at the same time, and set the initial learning rate to 0.0016.

\begin{table}[h]\scriptsize%
\caption{ Quantitative results of F-scores (higher means better) on Tanks and Temples dataset} \label{Tank_results}
\begin{center}
% \resizebox{\textwidth}{12mm}{
\begin{tabular}{llllllllll}
\hline
methods	&Family	&Francis &Horse	&Lighthouse	&M60 &Panther &Playground &Train &F-score\\
\hline
Colmap	&50.41	&22.25	&25.63	&56.43	&44.83	&46.97	&48.53	&42.04	&42.14\\
MVSNet	&55.99	&28.55	&25.07	&50.79	&53.96	&50.86	&47.90	&34.69	&43.48\\
R-MVSNet	&69.96	&46.65	&32.59	&42.95	&51.88	&48.80	&52.00	&42.38	&48.40\\
CVP-MVSNet  &\textbf{76.5}   &47.74  &36.34  &55.12  &\textbf{57.28}  &54.28  &57.43  &47.54  &54.03\\
Point-MVSNet	&61.79	&41.15	&34.20	&50.79	&51.97	&50.85	&52.38 &43.06	&48.27\\
La-Net &76.24 &54.32 &\textbf{49.85} &54.03 &56.08 &50.82 &53.71 &50.57 &55.70\\
Att-MVS &73.90 &\textbf{62.58} &44.08 &\textbf{64.88} &56.08 &\textbf{59.39} &\textbf{63.42} &\textbf{56.06} &\textbf{60.05}\\
UCSNet	&76.09	&53.16	&43.03	&54.00	&55.60	&51.49	&57.38	&47.89	&54.83\\
\hline
ours	&73.92	&\underline{53.88}	&\underline{44.84}	&56.06	&\underline{56.71}	&53.65	&\underline{59.83}	&48.86	&\underline{55.97}\\
\hline

\end{tabular}
\end{center}
\end{table}

\begin{figure*}[h]
\centering
    \subfigure[Ground-Truth]{\includegraphics[width=0.22\textwidth,height=0.45\textwidth]{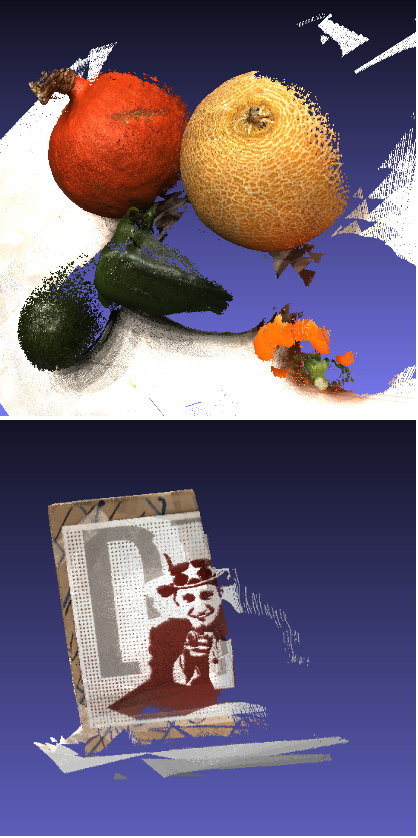}}
    \subfigure[UCSNet]{\includegraphics[width=0.22\textwidth,height=0.45\textwidth]{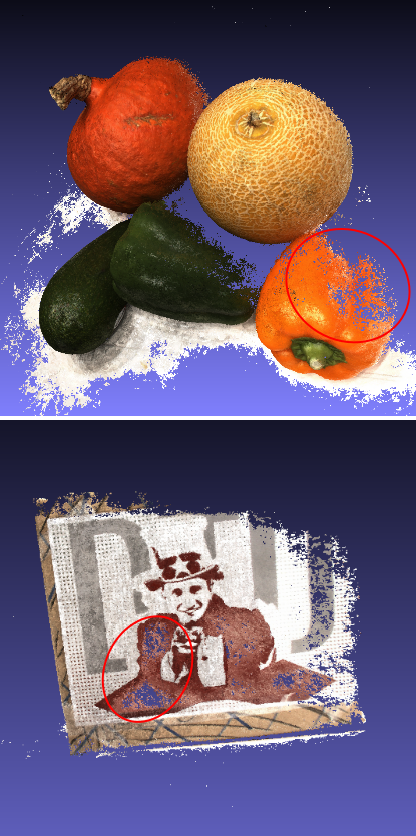}}
    \subfigure[Ours]{\includegraphics[width=0.22\textwidth,height=0.45\textwidth]{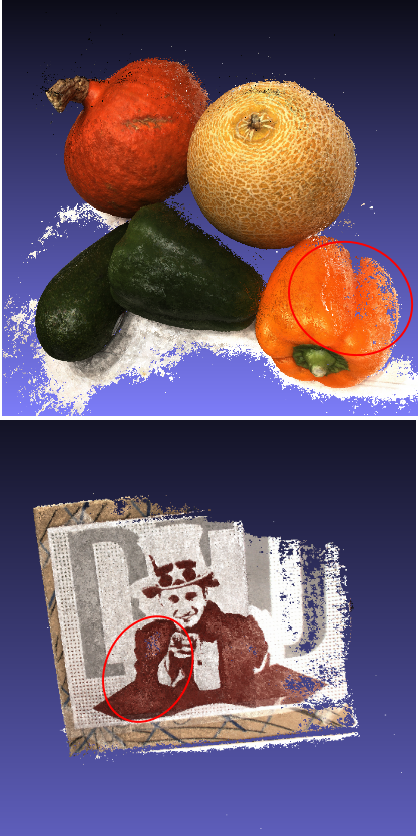}}
    \subfigure[Tank]{\includegraphics[width=0.22\textwidth,height=0.45\textwidth]{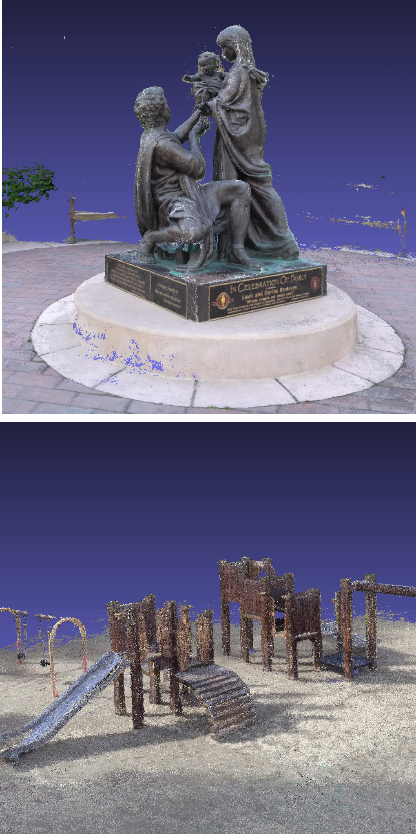}}
\caption{Qualitative analysis on DTU dataset and Tank\&Temples dataset. \textbf{(a) to (c)}: \textbf{Top Row:} Reconstruction on scan75. \textbf{Bottom Row:} Reconstruction on scan13. \textbf{(d)} shows the reconstruction effect of our network on the Tank\&Temples dataset.}
\label{dtu_figure}
\end{figure*}
% \begin{figure}
%     \centering
%     %\subfigure[Horse]{\includegraphics[width=4cm,height=4cm]{pics/horse.png}}
%     \subfigure[Family]{\includegraphics[width=4cm,height=4cm]{imgs/Family06.png}}
%     \subfigure[PlayGround]{\includegraphics[width=4cm,height=4cm]{imgs/playground00.png}}
%     \caption{Qualitative results of reconstructed point clouds on Tank and Temples benchmark.}
%     \label{Tank_figure}
% \end{figure}
\begin{table}[h]\scriptsize%
\caption{Ablation results on DTU testing set. ``MFA'' represents the multi-scale feature extractor with self-attention mechanism, ``AG'' is our attention thin volume aggregate process with the introduction of group-wise correlation. ``AT'' represents the combination of our feature extractor and our attention thin volume aggregate process. } \label{as_results}
\begin{center}
\begin{tabular}{l|ll|lll|lll|l}

\hline
% \multicolumn{2}{c}{\multirow{2}{*}{Model}} &\multirow{2}{*}{saUNet}&\multirow{2}{*}{group} & \multicolumn{3}{c}{Multi-metric Loss}& \multirow{2}{*}{Acc.} &\multirow{2}{*}{Comp.}& \multirow{2}{*}{Overall}\\
% \cline{6-8}

% \multicolumn{2}{|c|}{} & & & FL & NBL & DL  \\
\hline

method & MFA & AG & PL & NBL & DL & Acc. & Comp. & Overall & Memory\\
\hline
UCSNet	& & & & & \checkmark &\textbf{0.338}	&0.349	&0.344 &6226Mb\\
UCSNet+MFA &\checkmark & & & & \checkmark	&0.350	&0.327	&0.338 &7455Mb\\
UCSNet+AG & &\checkmark & & & \checkmark	&0.363	&0.316	&0.340 &5881Mb\\
UCSNet+AT &\checkmark &\checkmark & & & \checkmark	&0.357	&\underline{0.312}	&0.335 &6686Mb\\
\hline
UCSNet+AT+PL &\checkmark &\checkmark &\checkmark & & \checkmark	&0.350	&0.333	&0.341 &\\
UCSNet+AT+NBL &\checkmark &\checkmark & &\checkmark & \checkmark	&0.355	&\textbf{0.311}	&\underline{0.333} &\\
UCSNet+AT+PL+NBL &\checkmark &\checkmark &\checkmark &\checkmark & \checkmark	&\underline{0.347}	&0.315	&\textbf{0.331} &\\
\hline

\hline
\end{tabular}
\end{center}
\end{table}
% \begin{table}[h]
% \caption{Comparison of memory consumption and computational efficiency of different components. The size of input images is 640*512 pixels. All these experiments are run on a Tesla V100 GPU.} \label{memory_result}
% \begin{center}
% \begin{tabular}{p{0.2\textwidth}|p{0.125\textwidth}p{0.15\textwidth}|p{0.125\textwidth}p{0.15\textwidth}|p{0.125\textwidth}p{0.15\textwidth}}
% \hline
% Method & Mem(M)  & Comparison & Training Time(s/iter) &  Comparison & Inference Time(s/frame) & Comparison\\
% \hline
% baseline	&6226	 &- &1.058 &- &0.59 &- \\
% baseline+AG &5881 &-5.5\% &1 &-5.5\% &0.54 &-8.4\%\\
% baseline+MFA	&7455 &+19.7\% &1.203 &+13.7\% &0.87 &+47.5\%\\
% baseline+AT	&6686	&+7.4\% &1.152 &+8.9\% &0.81 & +37.3\%\\
% \hline
% \end{tabular}
% \end{center}
% \end{table}
\subsection{Benchmark Performance}\label{4.3}
\subsubsection{Overall evaluation on DTU dataset}
We evaluate our network on DTU testing set. For fair comparisons, we use the same view selection, image resolution and initial depth range as $N=5,H \times W=1600 \times 1184,d_{min}=425mm,d_{max}=933.8mm$.

The official Matlab evaluation scripts of DTU dataset is used to evaluate the accuracy and completeness of our results. It compares the distance between our produced point clouds and the ground-truth point clouds. Then we compare the results to state-of-the-art methods and show them in Table ~\ref{DTU_results}. 
It is not difficult to see that the traditional method Gipuma ranks first in accuracy metric, however, our approach performs better in \textbf{completeness} and \textbf{overall scores}(lower is better in both metrics). On the basis of quantitative analysis, we demonstrate the superiority of our method by visually comparing the reconstruction effects (see Figure~\ref{dtu_figure}). Our method can obtain a more complete surface point cloud at some positions where the past methods is more disadvantageous. 

% At the same time, it is worth noting that thanks to the coarse to fine strategy, the network can directly output the depth map of the original resolution. The original MVSNet and R-MVSNet can only get results that are one-fourth the size of the original resolution. This strategy provides favorable conditions for the reconstruction of large-scale scenes, which can be seen in the following section.

\subsubsection{Overall evaluation on Tank\&Temples dataset}
Now we show the generalization ability of our network. Without fine-tuning, we use the model trained on DTU dataset to test on Tank\&Temples intermediate dataset. During testing, we set the number of input images and the image resolution to $N=5,H=1920,W=1056$. The quantitative results are shown in the Table~\ref{Tank_results}. 

Our method performs better than the baseline(UCSNet~\cite{UCSNet}), achieving an average F-score (55.97), which is competitive in all public results. In addition, we have observed that in some difficult scenes, such as sand, cylindrical railings, etc., we can get a very complete point cloud, as shown in the Figure~\ref{dtu_figure}.
\subsection{Ablation Studies}\label{4.4}
In this section, ablation experiments are performed to evaluate the advantages of our work. The ablation experiments are performed and evaluated on the DTU dataset. We compare the baseline model UCSNet~\cite{UCSNet} with our approach on each component. This section will show the impact of each component. The results are in the Table~\ref{as_results}.\\
\textbf{Benefits of each modules} It is not difficult to find that both the self-attention module and attention thin volume aggregate module can improve the completeness. When the two components are superimposed, the improvement of the completeness score is the largest, while maintaining the high level of overall evaluation.\\
\textbf{Benefits of multi-metric loss function} As shown in Table~\ref{as_results}, applying multi-metric loss function can significantly improve the effect of reconstruction. The position loss term(PL) can improve the accuracy, while the fault tolerance term (NBL) is beneficial to the completeness metric. In qualitative analysis, we found that the improvement in completeness bought by NBL is mainly concentrated on the local scale, which is consistent with our assumption. When the two are combined, we can achieve a balance of two metrics for the best overall score.\\
\textbf{Efficiency discussion} We also pay attention to the impact of each module on GPU memory. The details are shown in Table \ref{as_results}. We can see that the memory usage increases significantly when the self-attention module is applied, and as we expected, the group-wise correlation strategy can reduce the memory consumption, so that the overall network size is not excessively increased. The combination of two components can reduce the dependence on memory as around 10\% under the premise of ensuring the reconstruction quality, which provides the possibility for large-scale scene reconstruction.

\section{Conclusion}\label{sec6}
In this paper, around building an efficient and practical cost volume, we analyze the association of neighboring pixels in the MVS problem, introduce the self-attention mechanism to design a multi-scale feature extractor, and then employ feature-wise loss function to enhance the spatial similarity of the corresponding feature vectors. The entire network follows the coarse to fine framework, combining group-wise correlation and uncertainty estimation strategies to construct a lightweight attention thin volume. With this help, our network can better express the geometric correspondence. After extensive evaluation of two challenging benchmark datasets, the results show that our approach is very competitive compared with other methods. In the future, we hope to further improve our network and make the learning based MVS method more efficient and reliable.

\end{document}